\documentclass[10pt,letterpaper]{article}

\usepackage{cogsci}
\cogscifinalcopy 

\usepackage{url}            
\usepackage{booktabs}       
\usepackage{amsfonts}       
\usepackage{nicefrac}       
\usepackage{microtype}      

\usepackage{pslatex}
\usepackage[utf8]{inputenc}
\usepackage{graphicx}
\usepackage{amsmath}
\usepackage{bm}
\usepackage{subcaption}
\usepackage{placeins}
\usepackage{wrapfig}
\usepackage[font=small,labelfont=bf]{caption}
\usepackage[english]{babel}
\usepackage{sidecap}
\usepackage{enumitem}
\usepackage{xcolor}
\usepackage{nidanfloat}
\usepackage[natbibapa]{apacite}
\usepackage{natbib}

\makeatletter
\newcommand\bibsize{\@setfontsize\bibsize{7.5}{8.5}}
\makeatother

\def \P{\mathrm{P}}
\def \E{\mathrm{E}}

\usepackage[pagebackref=true,breaklinks=true,bookmarks=false,hidelinks]{hyperref}

\title{Investigating Simple Object Representations in Model-Free Deep Reinforcement Learning}

\author{{\bf Guy Davidson (guy.davidson@nyu.edu)} \\
  Center for Data Science \\
  New York University
  \And {\bf Brenden M. Lake (brenden@nyu.edu)} \\
  Department of Psychology and Center for Data Science \\
  New York University}

\begin{document}

\maketitle

\begin{abstract}
We explore the benefits of augmenting state-of-the-art model-free deep reinforcement learning with simple object representations.
Following the Frostbite challenge posited by \cite{Lake2017}, we identify object representations as a critical cognitive capacity lacking from current reinforcement learning agents.
We discover that providing the Rainbow model \citep{Hessel2018} with simple, feature-engineered object representations substantially boosts its performance on the Frostbite game from Atari 2600.
We then analyze the relative contributions of the representations of different types of objects, identify environment states where these representations are most impactful, and examine how these representations aid in generalizing to novel situations.


\end{abstract}

Deep reinforcement learning (deep RL) offers a successful example of the interplay between cognitive science and machine learning research.
Combining neural network-based function approximation with psychologically-driven reinforcement learning, research in deep RL has achieved superhuman performance on Atari games \citep{Mnih2015}, board games such as chess and Go \citep{Silver2017}, and modern video games such as DotA 2 \citep{OpenAI2018}.
While the successes of deep RL are impressive, its current limitations are becoming increasingly more apparent. 
The algorithms require orders of magnitude more training to learn how to play such games compared to humans.
\cite{Tsividis2017} show that fifteen minutes allow humans to rival models trained for over 100 hours of human experience, and to an extreme, the OpenAI Five models \citep{OpenAI2018} collect around 900 years of human experience \textit{every day} they are trained.
Beyond requiring unrealistic quantities of experience, the solutions these algorithms learn are highly sensitive to minor design choices and random seeds \citep{Henderson2018} and often struggle to generalize beyond their training environment \citep{Cobbe2019,Packer2018}. 
We believe this suggests that people and deep RL algorithms are learning different kinds of knowledge, and using different kinds of learning algorithms, in order to master new tasks like playing Atari games. Moreover, the efficiency and flexibility of human learning suggest that cognitive science has much more to contribute to the development of new RL approaches.


In this work, we focus on the representation of objects as a critical cognitive ingredient toward improving the performance of deep reinforcement learning agents. Within a few months of birth, infants demonstrate sophisticated expectations about the behavior of objects, including that objects persist, maintain size and shape, move along smooth paths, and do not pass through one another \citep{Spelke1990,Spelke1992}.
In artificial intelligence research, \citet{Lake2017} point to object representations (as a component of intuitive physics) as an opportunity to bridge the gap between human and machine reasoning.
\citet{Diuk2008} utilize this notion to reformulate the Markov Decision Process (MDP; see below) in terms of objects and interactions, and \citet{Kasnky2017} offer Schema Networks as a method of reasoning over and planning with such object entities.
\citet{Dubey2018} explicitly examine the importance of the visual object prior to human and artificial agents, discovering that the human agents exhibit strong reliance on the objectness of the environment, while deep RL agents suffer no penalty when it is removed. 
More recently, this inductive bias served as a source of inspiration for several recent advancements in deep learning, such as  \citet{Kulkarni2016}, \citet{vanSteenkiste2018},  \citet{Kulkarni2019}, \citet{Veerapaneni2019}, and \citet{Lin2020}.
This work aims to contribute toward the Frostbite Challenge posited by \citet{Lake2017}, of building algorithms that learn to play Frostbite with human-like flexibility and limited training experience.
We aim to pinpoint the contribution of object representations to the performance of deep RL agents, in order to reinforce the argument for object representations (and by extension, other cognitive constructs) to such models.

To that end, we begin from Rainbow \citep{Hessel2018}, a state of the art model-free deep RL algorithm, and augment its input (screen frames) with additional channels, each a mask marking the locations of unique semantic object types. 
We confine our current investigation to the game of Frostbite, which allows crafting such masks from the pixel colors and locations in the image. 
While this potentially limits the scope of our results, it allows us to perform intricate analyses of the effects of introducing these object representations.
We do not consider the representations we introduce to be cognitively plausible models of the representations humans reason with; they are certainly overly simple.
Instead, our choice of representations offers a minimally invasive way to examine the efficacy of such representations within the context of model-free deep reinforcement learning. 
We find that introducing such representations aids the model in learning from a very early stage in training, surpassing in under 10M training steps the performance \citet{Hessel2018} report in 200M training steps.
We then report several analyses as to \textit{how} the models learn to utilize these representations: we examine how the models perform without access to particular channels, investigate variations in the value functions learned, and task the models with generalizing to novel situations.  
We find that adding such object representations tends to aid the models in identifying environment states and generalizing to a variety of novel situations, and we discuss several conditions in which these models are more (and less) successful in generalization. 

\section{Methodology}

\textbf{Reinforcement Learning.} 
Reinforcement learning provides a computational framework to model the interactions between an agent and an environment.
At each timestep $t$, the agent receives an observation $S_t \in \mathcal{S}$ from the environment and takes an action $A_t \in \mathcal{A}$, where $\mathcal{S}$ and $\mathcal{A}$ are the spaces of possible states and actions respectively. 
The agent then receives a scalar reward $R_{t+1} \in \mathbb{R}$ and the next observation $S_{t+1}$, and continues to act until the completion of the episode. 
Formally, we consider a \textit{Markov Decision Process} defined by a tuple $\langle \mathcal{S}, \mathcal{A}, T, R, \gamma \rangle$, where $T$ is the environment transition function, $T(s, a, s')= \P(S_{t+1} = s' | S_t = s, A_t = a)$, $R$ is the reward function, and $\gamma \in [0, 1]$ is a discount factor. 
The agent seeks to learn a policy $\pi(s, a) = \P (A_t = a | S_t = s)$ that maximizes the expected return $\E [G_t]$, where the return $G_t = \sum_{k=0}^{\infty} \gamma^k R_{t + k}$ is the sum of future discounted rewards. 
For further details, see \cite{SuttonBarto2018}. 

To construct a policy, reinforcement learning agents usually estimate one of two quantities.
One option is to estimate the state-value function: $v_\pi(s) = \E_\pi \left[ G_t | S_t = s \right]$. 
The state value reflects the expected discounted return under the policy $\pi$ starting at a state $s$.
Another oft-estimated quantity is the action-value (or Q-value) function: $q_\pi (s, a) = \E_\pi \left[ G_t | S_t = s, A_t = a \right]$, which models the expected discounted return from taking action $a$ at state $s$ (under policy $\pi$).
Note that the Q-values relate to the state values by $v_\pi(s) = \max_a q_\pi (s, a)$; the state value is equal to the action value of taking the optimal action.

\begin{figure}[!htb]
\centering
\includegraphics[width=\linewidth]{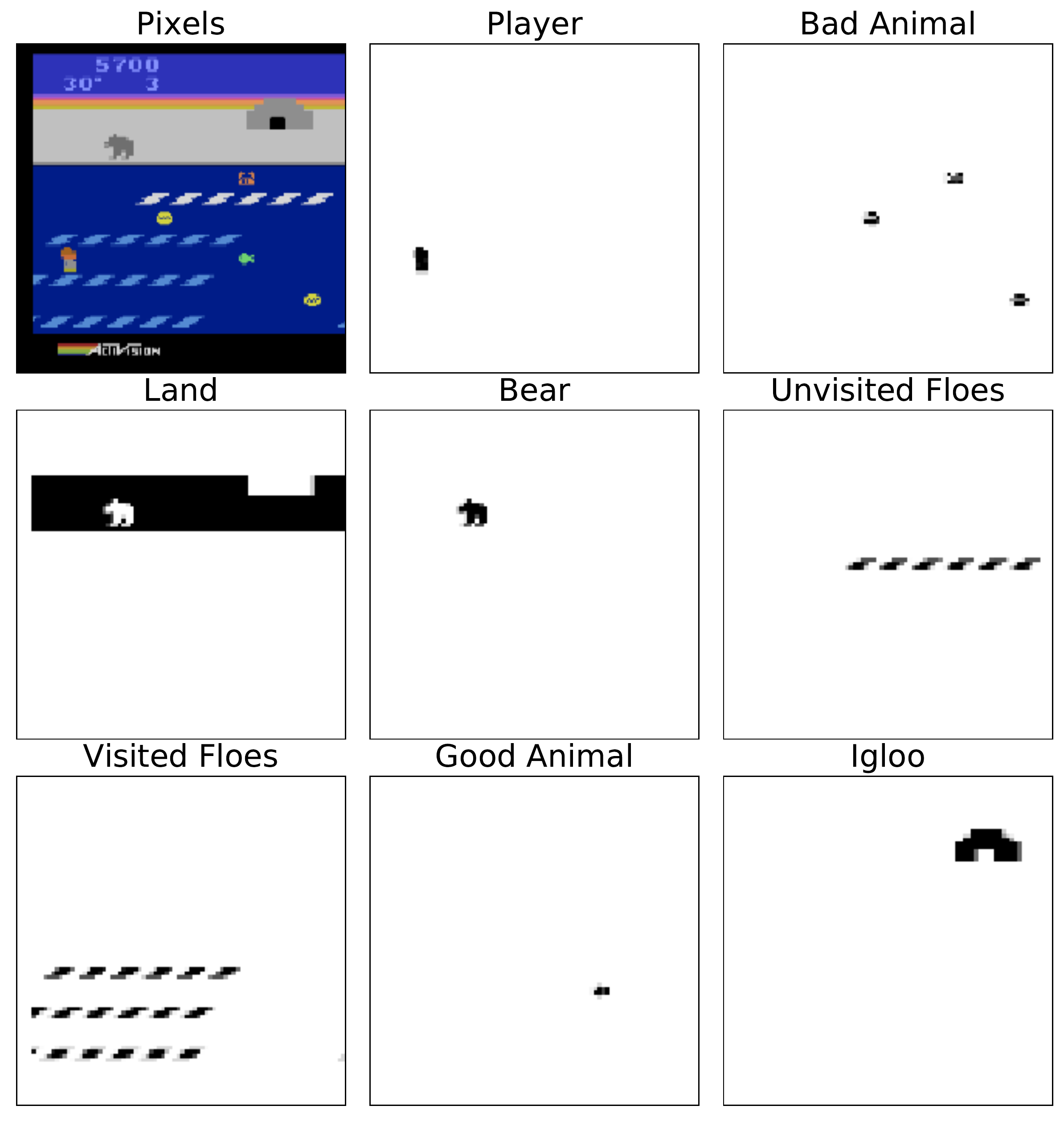}
\caption{ {\bf Object mask examples.} A sample screen frame encountered during Frostbite gameplay (top left), and the object masks computed from it (remaining panels).}
\label{fig:mask-examples}
\end{figure}

\textbf{Atari Learning Environment and Frostbite.} 
We utilize the Atari Learning Environment \citep{Bellemare2013,Machado2018}, and specifically the game Frostbite, as a testbed for our object representations. 
The ALE offers an interface to evaluate RL algorithms on Atari games and has served as an influential benchmark for reinforcement learning methods. 
In Frostbite\footnote{To play the game for yourself: \url{http://www.virtualatari.org/soft.php?soft=Frostbite}}
, the player controls an agent attempting to construct an igloo (top-right corner of pixels panel in \autoref{fig:mask-examples}) before they freeze to death when a time limit is hit.
To construct the igloo, the agent must jump between ice floes, making progress for each unvisited (in the current round) set of ice floes, colored white (compared to the blue of visited floes).
The ice floes move from one end of the screen to the other, each row in a different direction, in increasing speed as the player makes progress in the game.
As the agent navigates building the igloo, they must avoid malicious animals (which cause loss of life, in orange and yellow in \autoref{fig:mask-examples}), and may receive bonus points for eating fish (in green).
To succeed, the player (or model) must plan how to accomplish subgoals (visiting a row of ice floes; avoiding an animal; eating a fish) while keeping track of the underlying goal of completing the igloo and finishing the current level.

\textbf{DQN.}
As the Rainbow algorithm builds on DQN, we provide a brief introduction to the DQN algorithm, and refer the reader to \citet{Mnih2015} for additional details:
DQN is a model-free deep reinforcement learning algorithm: \textit{model-free} as it uses a parametric function to approximate Q-values, but does not construct an explicit model of the environment; \textit{deep} as it uses a deep neural network (see \cite{Goodfellow2016} for reference) to perform the approximation.
The model receives as state observations the last four screen frames (in lower resolution and grayscale), and emits the estimated Q-value for each action, using a neural network comprised of two convolutional layers followed by two fully connected layers. 
DQN collects experience by acting according to an $\epsilon$-greedy policy, using Q-values produced by current network weights (denoted $Q(s, a; \theta)$ for a state $s$, action $a$, and weights $\theta$). 
To update these weights, DQN minimizes the difference between the current Q-values predicted to bootstrapped estimates of the Q-value given the reward received and next state observed.

\textbf{Rainbow.} 
Rainbow \citep{Hessel2018} is an amalgamation of recent advancements in model-free deep reinforcement learning, all improving on the DQN algorithm. 
Rainbow combines the basic structure of DQN with several improvements, such as Prioritized Experience Replay \citep{Schaul2016}, offering a better utilization of experience replay; Dueling Networks \citep{Wang2016}, improving the Q-value approximation; and Distributional RL \citep{Bellemare2017}, representing uncertainty by approximating distributions of Q-values, rather than directly estimating expected values.
For the full list and additional details, see \cite{Hessel2018}. 
Note that save for a minor modification of the first convolutional layer, to allow for additional channels in the state observation, we make no changes to the network structure, nor do we perform any hyperparameter optimization.

\textbf{Object Segmentation Masks.} 
We compute our object segmentation masks from the full-resolution and color frames. 
Each semantic category (see \autoref{fig:mask-examples} for categories and example object masks) of objects in Frostbite is uniquely determined by its colors and a subset of the screen it may appear in. 
Each mask is computed by comparing each pixel to the appropriate color(s), and black or white-listing according to locations. 
As the masks are created in full-resolution, we then reduce them to the same lower resolution used for the frame pixels. 
We pass these masks to the model as separate input channels, just as Rainbow receives the four screen frame pixels.

\textbf{Experimental Conditions.} 
\textbf{Pixels:} In this baseline condition, we utilize Rainbow precisely as in \cite{Hessel2018}, save for a minor change in evaluation described below. 
\textbf{Pixels+Objects:} 
In this condition, the model receives both the four most recent screen frames, as well as the eight object masks for each frame.
This requires changing the dimensionality of the first convolutional layer, slightly increasing the number of weights learned, but otherwise the model remains unchanged.
\textbf{Objects:} 
We wanted to understand how much information the screen pixels contain beyond the object masks we compute.
In this condition, we omit passing the screen frames, instead passing only the masks computed from each frame. 
\textbf{Grouped-Moving-Objects:} We wished to investigate the importance of separating objects to separate semantic channels. 
Loosely inspired by evidence that animacy influences human object organization \citep{Konkle2013}, we combined all object masks for moving objects (all except for the land and the igloo) to a single binary mask, which we passed the model alongside the state pixels.  
For the remainder of this paper, we shall refer to this condition as \textit{Grouped}.

\textbf{Training and evaluation.} 
Each model is trained for ten million steps, or approximately 185 human hours\footnote{At 60 frames per second, one human hour is equivalent to 216,000 frames, or 54,000 steps, as each step includes four new frames. See \cite{Mnih2015} for additional details.} identically to the models reported in \cite{Hessel2018}. 
Our evaluation protocol is similar to \textit{no-op starts} protocol used by \cite{Hessel2018}, except that rather than evaluating models for 500K frames (and truncate episodes at 108K), we allow each model ten evaluation attempts without truncation (as Frostbite as a relatively challenging game, models rarely exceed 20K frames in a single episode. 
We perform ten replications of the baseline \textit{Pixels} condition, recovering similar results to those reported by \cite{Hessel2018}, and ten replications of each of our experimental conditions.

\textbf{Implementation.} 
All models were implemented in PyTorch \citep{Paszke2017}. Our modifications of Rainbow built  on Github user @Kaixhin's implementation of the Rainbow algorithm\footnote{\url{https://github.com/Kaixhin/Rainbow}}.

\section{Results}
\begin{figure}[!htb]
\centering
\includegraphics[width=\linewidth]{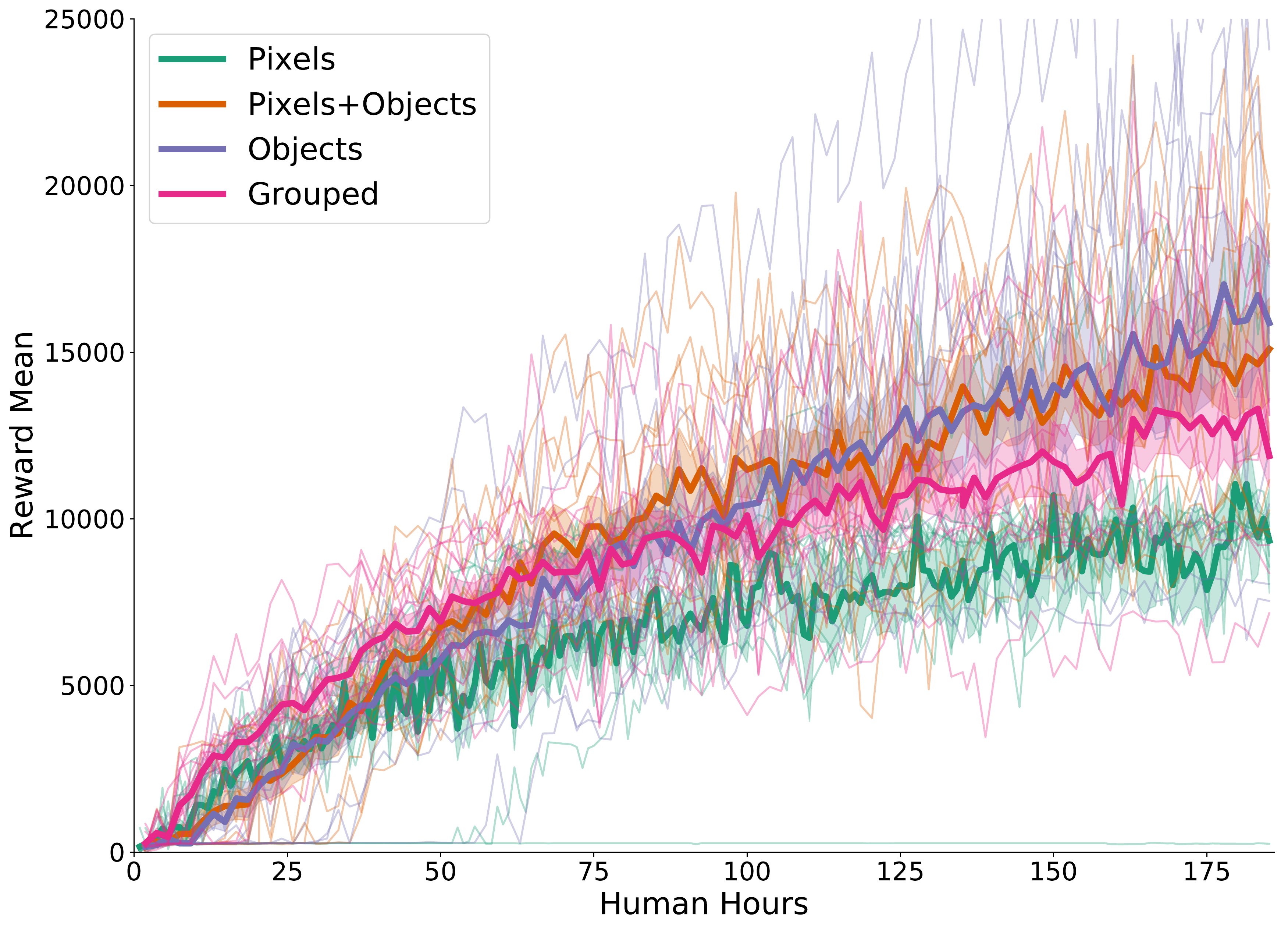}
\caption{ {\bf Mean evaluation reward per condition.} Mean evaluation reward per condition. Bold lines plot the mean per condition. Shaded areas reflect the standard error of the mean. Thin lines mark individual replications.}
\label{fig:mean-results}
\end{figure}

\autoref{fig:mean-results} summarizes the learning curves, depicting the mean evaluation reward received in each experimental condition over the first 10M frames (approximately 185 human hours) of training (see results for each human day of training in \autoref{table:mean-results}). 
Both conditions receiving the object masks show markedly better performance, breaking past the barrier of approximately 10,000 points reported by state-of-the-art model-free deep RL methods (including \cite{Hessel2018}), and continuing to improve. 
We also validated that this improvement cannot be explained solely by the number of parameters.\footnote{We evaluated a variant of the \textit{Pixels} model with additional convolutional filters, to approximately match the number of parameters the \textit{Pixels+Objects} has. 
The results were effectively identical to the baseline \textit{Pixels} ones.}
We find that including the semantic information afforded by channel separation offers a further benefit to performance beyond grouping all masks together, as evidenced by the \textit{Grouped} condition models falling between the baseline \textit{Pixels} ones and the ones with object information.
We find these results promising: introducing semantically meaningful object representations enable better performance in a fraction of the 200M frames that \cite{Hessel2018} train their models for.
However, the results for the first 24 human hours (see \autoref{fig:mean-results-first-day} in the Appendix) indicate that initial learning performance is not hastened, save for some advantage provided by the \textit{Grouped} models over the remainder.
One interpretation of this effect is that the single channel with moving objects best mirrors early developmental capacities for segmenting objects based on motion \citep{Spelke1990}.
We take this as evidence that object representations alone will not bridge the gap to human-like learning. 

We are less concerned with the improvement per se and more with what these object representations enable the model to learn, and how its learning differs from the baseline model. 
To that end, we dedicate the remainder of the results section to discussing several analyses that attempt to uncover how the models learn to utilize the object masks and what benefits they confer.

\begin{table}
\fontsize{7}{8}
\begin{tabular}{lllll}
\hline
 \textbf{Training}   & \textbf{Pixels}          & \textbf{Pixels+Objects}   & \textbf{Objects}   & \textbf{Grouped}       \\
\hline
 1 days     & $2443 \pm 661$  & $2352 \pm 617$   & $2428 \pm 612$   & $4426 \pm 585$   \\
 2 days     & $4636 \pm 792$  & $6220 \pm 618$   & $5366 \pm 863$   & $7318 \pm 560$   \\
 3 days     & $6460 \pm 1132$ & $8915 \pm 730$   & $7612 \pm 904$   & $8417 \pm 627$   \\
 4 days     & $7102 \pm 1027$ & $10066 \pm 876$  & $9761 \pm 1028$  & $9688 \pm 614$   \\
 5 days     & $8011 \pm 1112$ & $11239 \pm 1301$ & $11677 \pm 1559$ & $10111 \pm 843$  \\
 6 days     & $8236 \pm 1532$ & $13403 \pm 1157$ & $13033 \pm 1613$ & $11570 \pm 927$  \\
 7 days     & $9748 \pm 1652$ & $14277 \pm 1282$ & $14698 \pm 2038$ & $13169 \pm 1209$ \\
\hline
\end{tabular}
\caption{ {\bf  Mean evaluation results.} 
We report mean evaluation results for each condition after each day of human experience (approximately 1.3M training steps\protect\footnotemark[2]). 
Errors reflect the standard error of the mean. }
\label{table:mean-results}
\end{table}

\subsection{Omitted Objects Analysis}

\begin{figure}[t]
\centering
\includegraphics[width=\linewidth]{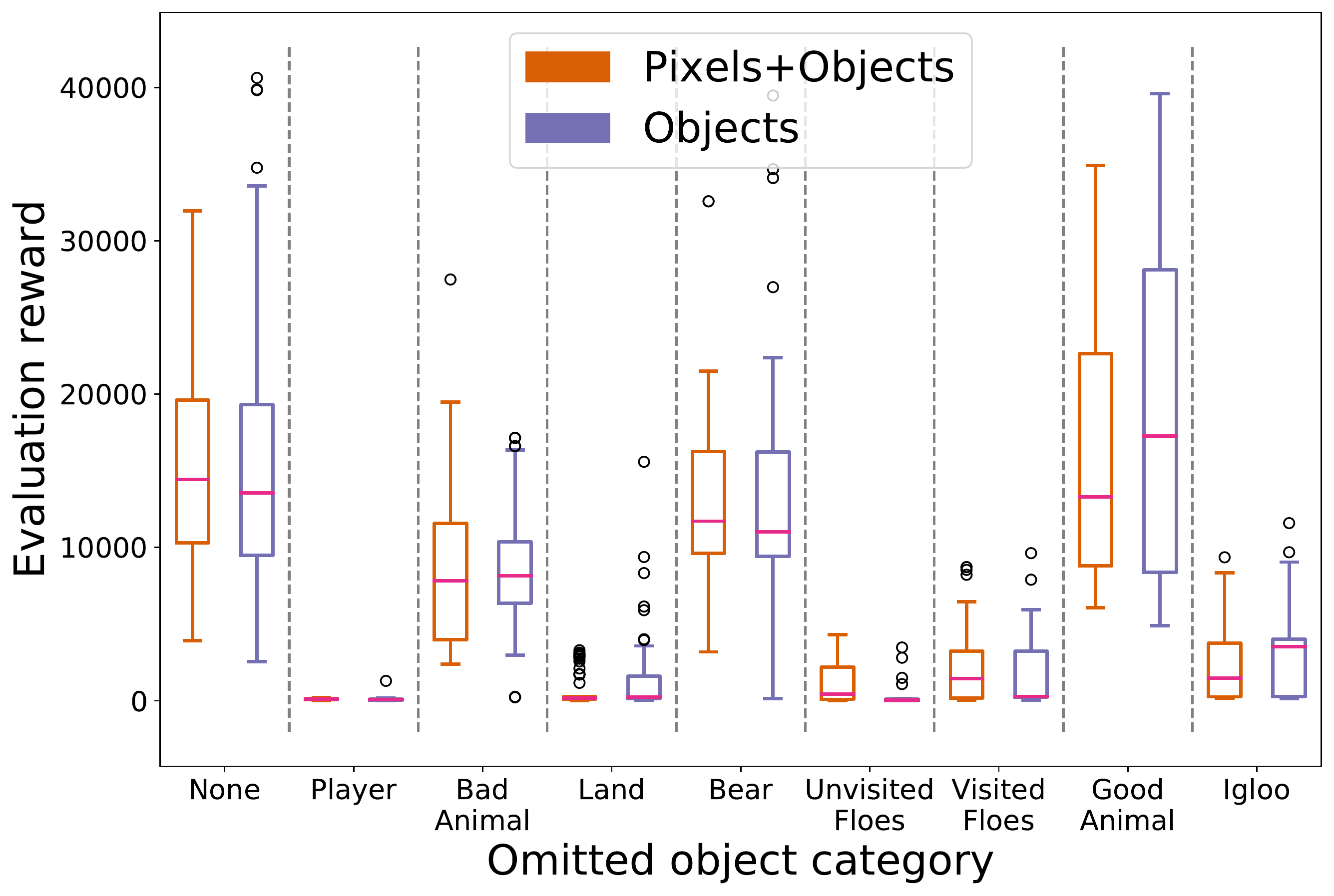}
\caption{ {\bf Omitted object comparison.} 
Comparing the resilience of models in the \textit{Pixels+Objects} and \textit{Objects} conditions to omitting specific object masks.
Each pair of box plots represents omitting a particular semantic category, with the full (no objects omitted) on the far left.
Pink lines mark the median of each distribution, boxes the first and third quartiles, whiskers to any points with 1.5 the interquartile range, and circles to any outliers.}
\label{fig:omited-objects}
\end{figure}

Our first analysis investigates the value of each type of object mask. 
To do so, we took models trained with access to all masks and evaluated them with one mask at a time omitted.
As to not interfere with input dimensionality, we omitted each mask by zeroing it out. 
In each experimental condition (\textit{Pixels+Objects} and \textit{Objects}), we evaluated each replication ten times without each mask. 
The boxplots in \autoref{fig:omited-objects} mark the distributions of evaluation rewards obtained across the different replications and evaluations. 
With minor exceptions, we find that performance is similar in the two conditions. 
It appears that while the \textit{Pixels+Objects} model receives the pixels, it primarily reasons based on the objects, and therefore sees its performance decline similarly to the \textit{Objects} models when the object masks are no longer available.

We find that some of the object masks are much more crucial than others: without the representations identifying the agent, the land, the ice floes, and the igloo (which is used to complete a level), our models' performance drops tremendously. 
The two representations marking antagonist animals (`bad animals' and bear) also contribute to success, albeit on a far smaller magnitude. 
Omitting the mask for the bear induces a smaller penalty as the bear is introduced in a later game state than other malicious animals. 
Finally, omitting the mask for the good (edible for bonus points) animals increases the variability in evaluation results, and, and in the \textit{Objects} condition, results in a higher median score. 
We interpret this as a conflict between short-term reward (eating an animal) and long-term success (finishing a level and moving on to the next state). 
Omitting information about the edible animals eliminates this tradeoff, and allows the model to `focus' on finishing levels to accrue reward.

These results indicate that models with both pixels and object representations lean on the object representations, and do not develop any redundancy with the pixels.
This analysis also validates the notion that the most important objects are the ones that guide the model to success, and that some objects may be unhelpful even when directly predictive of reward. 

\subsection{State-value difference comparison}

\begin{figure*}[!htb]
\vspace{-0.15in}
\centering
\includegraphics[width=\linewidth]{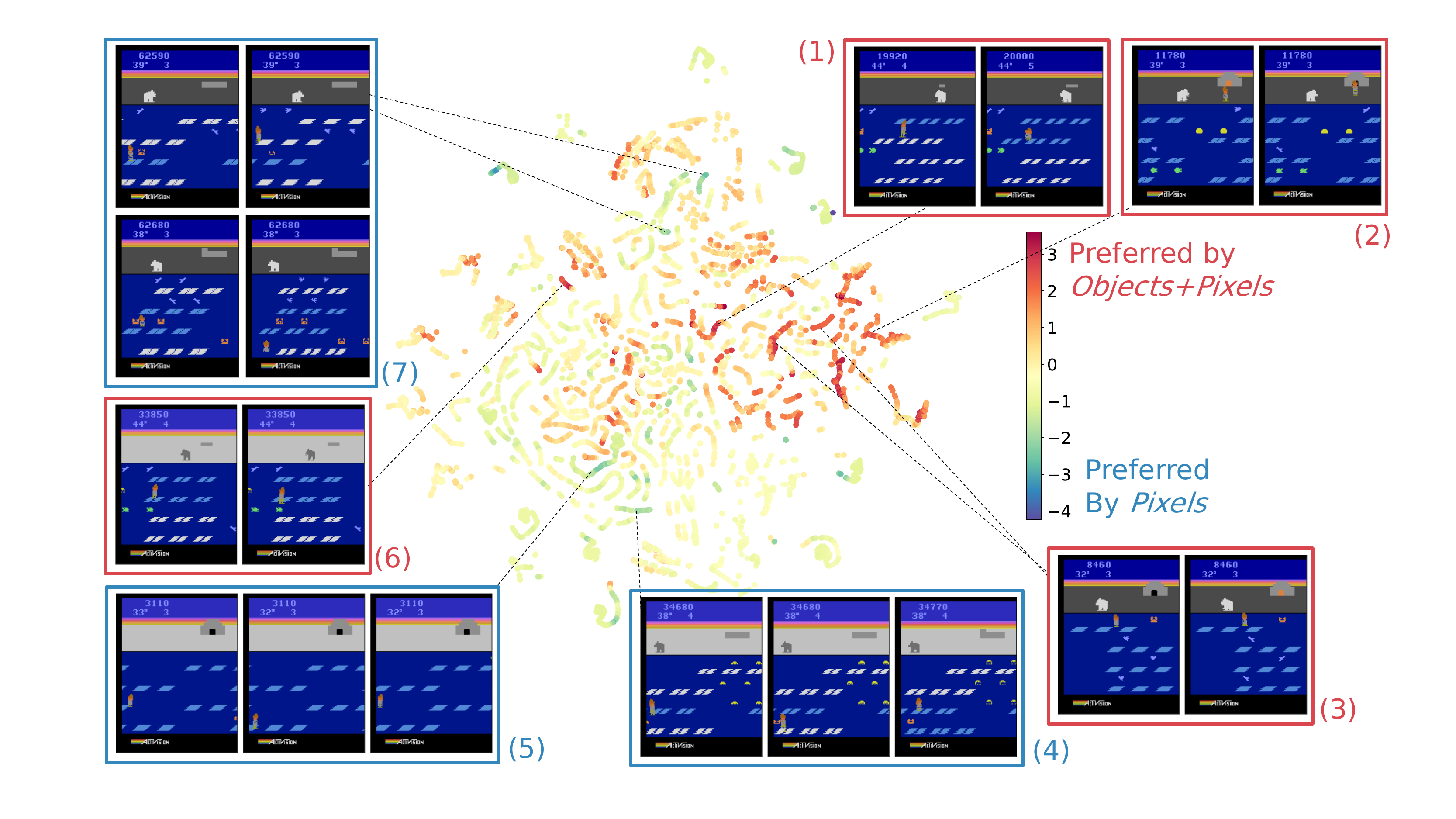}
\caption{ {\bf t-SNE visualization of internal model representations.} 
We use a single \textit{Pixels} condition model to compute latent representations of states collected during an evaluation, and embed them in 2D using t-SNE.
We color the embeddings by the normalized difference in mean state-values \protect\footnotemark\ between models in the \textit{Pixels+Objects} condition and models in the \textit{Pixels} condition.
Positive differences (in red) indicate higher state values in \textit{Pixels+Objects} models, negative differences (in blue) are states valued higher by \textit{Pixels} models. 
Sample states plotted (labeled 1-7 clockwise) represent observed clusters where object representations most affect the learned value functions. 
Border colors denote which condition predicted higher state values.}
\label{fig:tsne}
\end{figure*}

Next we highlight game states such that the object-augmented networks predict substantially different state values than networks based on pixels, adapting the t-SNE \citep{VanDerMaaten2008} methodology (using an implementation by \cite{Ulyanov2016}) utilized by \cite{Mnih2015} to visualize state representations.
We report a single comparison, between the replications in the baseline \textit{Pixels} condition and the replications in the \textit{Pixels+Objects} condition.
We collected sample evaluation states from the most successful (by evaluation scores) random seed of our \textit{Pixels} condition models. 
Utilizing the identity $v_\pi(s) = \max_a q_\pi (s, a)$, we passed each state through each of the models compared and took the maximal action value as the state value. 
We then colored the embeddings plotted in \autoref{fig:tsne} using the (normalized) difference in means, and examined clusters of states with extreme differences to understand differences in what the models succeeded in learning.
We plot the screen frames corresponding to these telling states on the circumference of \autoref{fig:tsne}.

States (2) and (3) highlight one failure mode of the \textit{Pixels} model: failure to recognize impending successful level completion.
The baseline \textit{Pixels} models miss the completed igloo (which terminates a level) on the dark background, and therefore fail to identify these states as highly rewarding -- even though the igloo itself remains identical to previous levels--the sole change, which \textit{Pixels} models fail to generalize to, is the change in background color.
Note that the \textit{Pixels} models do encounter the igloo on a dark background during training, but fail to recognize it--which appears to contribute to the score difference between these models and the ones that receive objects.
States (1) and (6) pinpoint another source of confusion: at some stage of the game, the ice floes begin splitting and moving relative to each other, rather than only relative to the overall environment.
Unlike the previous case, the \textit{Pixels} models do not encounter the splitting floes during training, as they appear in the level following the dark background igloos of states (2) and (3). 
Perhaps with training on these states the \textit{Pixels} models would learn to act around them; but as it stands, these models fail to generalize to this condition. 
This motion appears to confuse the baseline \textit{Pixels} models; we conjecture that the object representations enable the \textit{Pixels+Objects} models to reason better about this condition. 

States (4), (5), and (7) depict situations where the baseline \textit{Pixels} models erroneously fail to notice dangerous situations.
For example, in state (5), although the igloo is complete, the \textit{Pixels+Objects} models identify that the agent appears to be stuck without a way to return to land.
Indeed, the agent successfully jumps down, but then cannot return up and jumps into the water (and to its death).
States (4) and (7) represent similarly dangerous environment states, which the model used to collect the data survives in state (4) and succumbs to in state (7)\footnote{Note that states (4) and (7), like states (1) and (6), feature the splitting ice floes; the baseline \textit{Pixels} models' failure to recognize those might contribute to the differences in states (4) and (7) as well.}.
In these states, the \textit{Pixels+Objects} models appear to identify the precarious situation the agent is in, while the \textit{Pixels} models without object representations do not. 

\footnotetext{We pass each state through the ten replications in each condition, compute the expected state value (expected as Rainbow utilizes distributional RL), average across each condition for each state, compute the difference of means for each state, and normalize the collection of differences.}

\subsection{Generalization to novel situations}

\begin{figure*}[!htb]
\centering
\includegraphics[width=\linewidth]{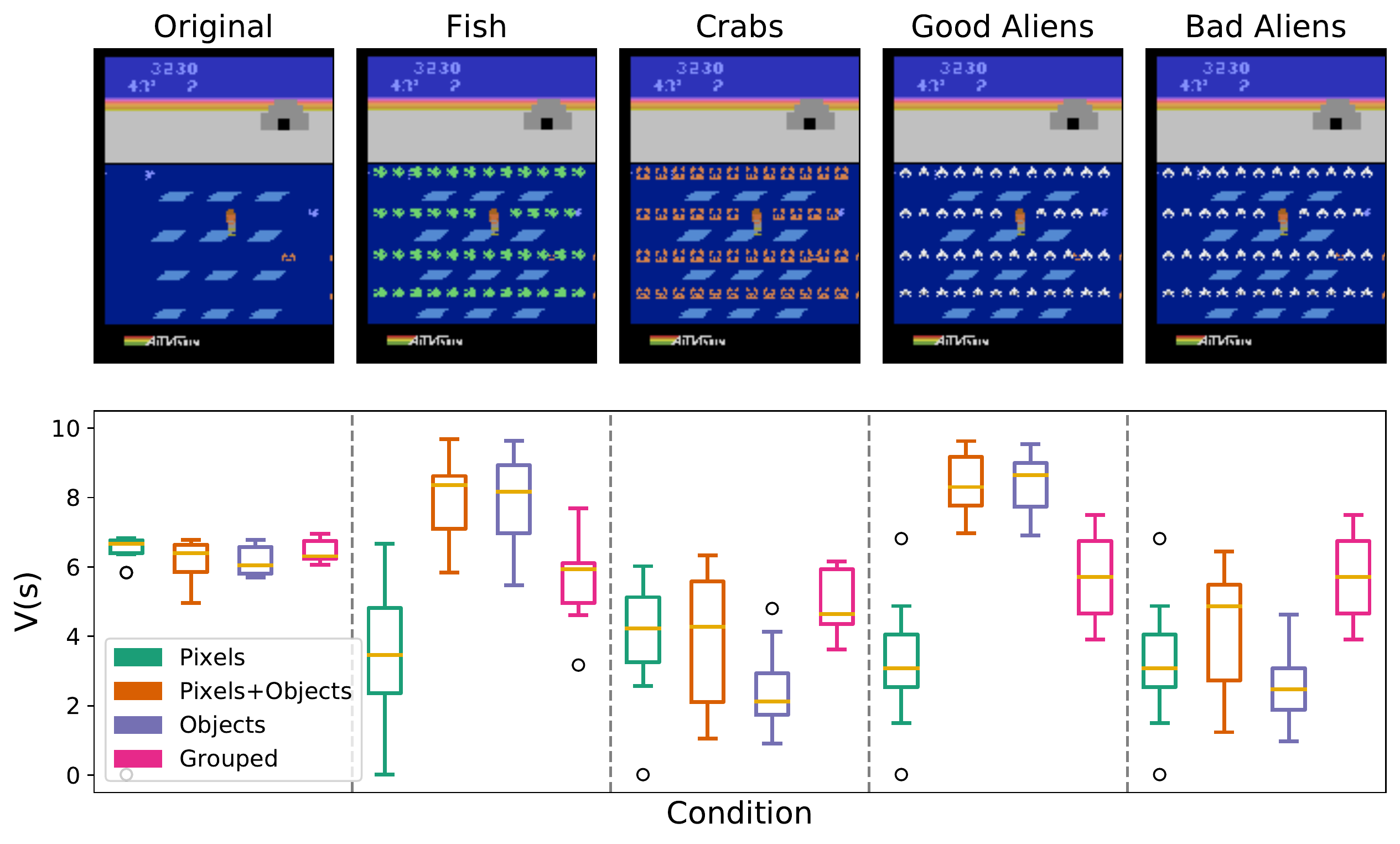}
\caption{ {\bf Generalization to being surrounded.} 
We examined how each model would react to being surrounded. 
In the first two manipulations, ``Fish'' and ``Crabs,'' we surround the agent with existing creatures from Frostbite.
In the second two, we borrow an alien sprite from the Atari game Space Invaders and label it in different object masks: first as edible fish, then as dangerous crabs.}
\label{fig:novel-aliens}
\end{figure*}

\begin{figure}[!htb]
\centering
\includegraphics[width=\linewidth]{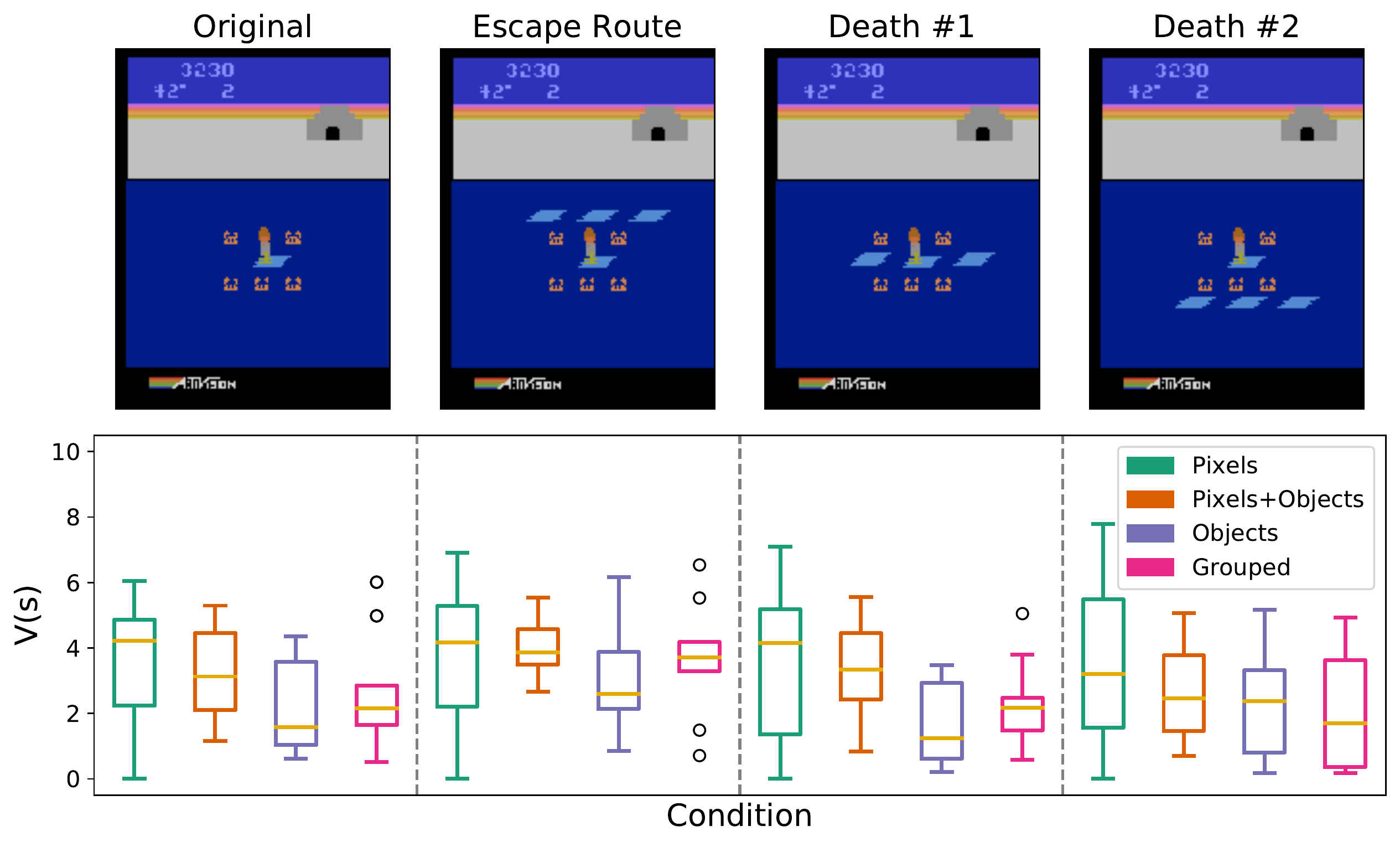}
\caption{ {\bf Generalization to being stranded among crabs.} 
We examined how each model would react when stranded and surrounded with crabs, with introducing additional ice floes that either serve as an escape route or a distraction.}
\label{fig:novel-stranded-crabs}
\end{figure}

We now investigate the utility of our object representations in generalizing to situations radically different from anything encountered during training, in hopes of investigating the capacity for systematic, human-like generalization.
In \autoref{fig:novel-aliens} and \autoref{fig:novel-stranded-crabs}, we augmented a state with additional features, adjusting both the pixels and object masks accordingly. 
We report a boxplot of state values predicted by models in each condition, both for the original state before any manipulation and for each manipulated state.

In \autoref{fig:novel-aliens}, we attempt to overwhelm the agent by surrounding it with animals--during normal gameplay, the model does not see anything resembling this simultaneous quantity of animals.
We also tested how our models would react to a novel object (an alien from Space Invaders), marked either as a `good animal' or as a `bad animal.'
The \textit{Pixels} models respond negatively to all of the overwhelming conditions, which is incorrect behavior in the Fish scene and correct behavior otherwise. 
Both \textit{Pixels+Objects} models and \textit{Objects} models show much stronger robustness to this change in distribution, reacting favorably to being surrounded with edible animals, and negatively to being surrounded with dangerous animals, demonstrating the generalization benefit conferred by our object representations.
Some of these differences in state-value distributions proved statistically significant: in a two-sample $t$-test, the differences in between the \textit{Pixels} and \textit{Pixels+Objects} models were significant ($P < 0.001$) in the `Fish' and `Good Aliens' condition. 
The same held true for the \textit{Pixels} and \textit{Objects} models. 
Comparisons between the \textit{Grouped} model and \textit{Pixels+Objects}  showed the same pattern of significance, in the `Fish' and `Good Aliens' cases, while the difference between the \textit{Grouped} and \textit{Objects} models proved significant ($P < 0.002$) in all four generalization scenarios.
When comparing within a model between the different manipulations, results were equally stark: for the \textit{Pixels+Objects} and \textit{Objects} models, \textit{all} differences were significant, save for the two with identical valence (`Fish' and `Good Aliens,' `Crabs' and `Bad Aliens').
Note that these are entirely novel objects to the agents, not merely unusual configurations of prior objects.
The entirely novel alien pixels appear to have little impact on the \textit{Pixels+Objects} models, providing further evidence that these models rely on the object masks to reason (rather than on the raw pixel observations).

We offer a second, potentially harder example, in which object representations do not seem sufficient to enable generalization. 
In \autoref{fig:novel-stranded-crabs}, we examine how the models respond to another situation far off the training data manifold: on a solitary ice floe, surrounded by crabs, with nowhere to go. 
Only in one scenario (`Escape Route') does the agent have a path to survival, and once arriving on land, could immediately finish the level and receive a substantial bonus. 
While some of the models show a preference to that state over the two guaranteed death states, the differences are fairly minor, and only one (for the \textit{Pixels+Objects} model, between `Death \#2' and `Escape Route') proves statistically significant.
This scenario, both its original state and the various manipulations performed, fall far off the data manifold, which may account for the higher variance in state values assigned to original state (compared to the one exhibited in \autoref{fig:novel-aliens}).

\section{Discussion}

We report a careful analysis of the effect of adding simple object masks to the deep reinforcement learning algorithm Rainbow evaluated on the Atari game Frostbite. 
We find that these representations enable the model to learn better and achieve higher overall scores in a fraction of the training time.
We then dissect the models to find the effect of these object representations.
We analyze their relative contributions to the model's success, we find states where object representations drastically change the learned value function, and we examine how these models would respond to novel scenarios. 
While we believe this provides strong preliminary evidence to the value of object representations for reinforcement learning, we acknowledge there is plenty of work ahead. 
We aim to investigate utilizing similar representations in other reinforcement learning tasks and explore methods towards learning robust, cognitively-plausible object representations, rather than providing them externally.
Furthermore, we believe that additional core ingredients of human cognition may prove as useful (if not more so) to the development of flexible and rapidly-learning reinforcement learning algorithms.
We hope to explore the benefits offered by an intuitive physical understanding of the world can confer, as it can allow an agent to predict the physical dynamics of the environment and plan accordingly.
We also believe that the capacity to develop causal models explaining observations in the environment, and the ability to meta-learn such models across different scenarios and games, would provide another step forward in the effort to build genuinely human-like reinforcement learning agents. 

\section{Acknowledgements}
BML is grateful for discussions with Tomer Ullman, Josh Tenenbaum, and Sam Gershman while writing ``Building machines that learn and think like people.'' 
Those ideas sit at the heart of this paper. 
We also thank Reuben Feinman and Wai Keen Vong for helpful comments on this manuscript.

\renewcommand*{\bibfont}{\bibsize}
\bibliographystyle{apalike}
\setlength{\bibleftmargin}{.125in}
\setlength{\bibindent}{-\bibleftmargin}
\bibliography{drl_object_representations} 

\begin{thebibliography}{}

\bibitem[Bellemare et~al., 2017]{Bellemare2017}
Bellemare, M.~G., Dabney, W., and Munos, R. (2017).
\newblock {A Distributional Perspective on Reinforcement Learning}.
\newblock In {\em Proceedings of the 34 th International Conference on Machine
  Learning, Sydney, Australia, PMLR 70}.

\bibitem[{Bellemare} et~al., 2013]{Bellemare2013}
{Bellemare}, M.~G., {Naddaf}, Y., {Veness}, J., and {Bowling}, M. (2013).
\newblock The arcade learning environment: An evaluation platform for general
  agents.
\newblock {\em Journal of Artificial Intelligence Research}, 47:253--279.

\bibitem[Cobbe et~al., 2019]{Cobbe2019}
Cobbe, K., Klimov, O., Hesse, C., Kim, T., and Schulman, J. (2019).
\newblock {Quantifying Generalization in Reinforcement Learning}.
\newblock In {\em Proceedings of the 36th International Conference on Machine
  Learning, Long Beach, California, PMLR 97}.

\bibitem[Diuk et~al., 2008]{Diuk2008}
Diuk, C., Cohen, A., and Littman, M.~L. (2008).
\newblock An object-oriented representation for efficient reinforcement
  learning.
\newblock In {\em Proceedings of the 25th International Conference on Machine
  Learning}, ICML ’08, page 240–247, New York, NY, USA. Association for
  Computing Machinery.

\bibitem[Dubey et~al., 2018]{Dubey2018}
Dubey, R., Agrawal, P., Pathak, D., Griffiths, T.~L., and Efros, A.~A. (2018).
\newblock Investigating human priors for playing video games.
\newblock In {\em ICML}.

\bibitem[Goodfellow et~al., 2016]{Goodfellow2016}
Goodfellow, I., Bengio, Y., and Courville, A. (2016).
\newblock {\em {Deep Learning}}.
\newblock MIT Press.

\bibitem[Henderson et~al., 2018]{Henderson2018}
Henderson, P., Islam, R., Bachman, P., Pineau, J., Precup, D., and Meger, D.
  (2018).
\newblock {Deep Reinforcement Learning that Matters}.
\newblock In {\em Thirthy-Second AAAI Conference On Artificial Intelligence
  (AAAI)}.

\bibitem[Hessel et~al., 2018]{Hessel2018}
Hessel, M., Modayil, J., van Hasselt, H., Schaul, T., Ostrovski, G., Dabney,
  W., Horgan, D., Piot, B., Azar, M., and Silver, D. (2018).
\newblock {Rainbow: Combining Improvements in Deep Reinforcement Learning}.
\newblock In {\em AAAI'18}.

\bibitem[Kansky et~al., 2017]{Kasnky2017}
Kansky, K., Silver, T., M\'{e}ly, D.~A., Eldawy, M., L\'{a}zaro-Gredilla, M.,
  Lou, X., Dorfman, N., Sidor, S., Phoenix, S., and George, D. (2017).
\newblock Schema networks: Zero-shot transfer with a generative causal model of
  intuitive physics.
\newblock In {\em Proceedings of the 34th International Conference on Machine
  Learning - Volume 70}, ICML’17, page 1809–1818. JMLR.org.

\bibitem[Konkle and Caramazza, 2013]{Konkle2013}
Konkle, T. and Caramazza, A. (2013).
\newblock {Tripartite organization of the ventral stream by animacy and object
  size}.
\newblock {\em Journal of Neuroscience}, 33(25):10235--10242.

\bibitem[Kulkarni et~al., 2019]{Kulkarni2019}
Kulkarni, T.~D., Gupta, A., Ionescu, C., Borgeaud, S., Reynolds, M., Zisserman,
  A., and Mnih, V. (2019).
\newblock {Unsupervised Learning of Object Keypoints for Perception and
  Control}.
\newblock In {\em Advances in Neural Information Processing Systems}, pages
  10723--10733.

\bibitem[Kulkarni et~al., 2016]{Kulkarni2016}
Kulkarni, T.~D., Narasimhan, K.~R., Saeedi, A., and Bcs, J. B.~T. (2016).
\newblock {Hierarchical Deep Reinforcement Learning: Integrating Temporal
  Abstraction and Intrinsic Motivation}.
\newblock In {\em Advances in Neural Information Processing Systems 29}, pages
  3675--3683.

\bibitem[Lake et~al., 2017]{Lake2017}
Lake, B.~M., Ullman, T.~D., Tenenbaum, J.~B., and Gershman, S.~J. (2017).
\newblock {Building machines that learn and think like people}.
\newblock {\em Behavioral and Brain Sciences}, 40(e253).

\bibitem[Lin et~al., 2020]{Lin2020}
Lin, Z., Wu, Y.-F., {Vishwanath Peri}, S., Sun, W., Singh, G., Deng, F., Jiang,
  J., and Ahn, S. (2020).
\newblock {SPACE: Unsupervised Object-Oriented Scene Representation via Spatial
  Attention and Decomposition}.
\newblock In {\em International Conference on Learning Representations}.

\bibitem[Machado et~al., 2018]{Machado2018}
Machado, M.~C., Bellemare, M.~G., Talvitie, E., Veness, J., Hausknecht, M.~J.,
  and Bowling, M. (2018).
\newblock Revisiting the arcade learning environment: Evaluation protocols and
  open problems for general agents.
\newblock {\em Journal of Artificial Intelligence Research}, 61:523--562.

\bibitem[Mnih et~al., 2015]{Mnih2015}
Mnih, V., Kavukcuoglu, K., Silver, D., Rusu, A.~A., Veness, J., Bellemare,
  M.~G., Graves, A., Riedmiller, M., Fidjeland, A.~K., Ostrovski, G., Petersen,
  S., Beattie, C., Sadik, A., Antonoglou, I., King, H., Kumaran, D., Wierstra,
  D., Legg, S., and Hassabis, D. (2015).
\newblock {Human-level control through deep reinforcement learning}.
\newblock {\em Nature}, 518(7540):529--533.

\bibitem[OpenAI, 2018]{OpenAI2018}
OpenAI (2018).
\newblock {OpenAI Five}.

\bibitem[Packer et~al., 2018]{Packer2018}
Packer, C., Gao, K., Kos, J., Kr{\"{a}}henb{\"{u}}hl, P., Koltun, V., and Song,
  D. (2018).
\newblock {Assessing Generalization in Deep Reinforcement Learning}.

\bibitem[Paszke et~al., 2017]{Paszke2017}
Paszke, A., Gross, S., Chintala, S., Chanan, G., Yang, E., DeVito, Z., Lin, Z.,
  Desmaison, A., Antiga, L., and Lerer, A. (2017).
\newblock {Automatic differentiation in PyTorch}.
\newblock In {\em NIPS}.

\bibitem[Schaul et~al., 2016]{Schaul2016}
Schaul, T., Quan, J., Antonoglou, I., and Silver, D. (2016).
\newblock {Prioritized Experience Replay}.
\newblock In {\em International Conference on Learning Representations}.

\bibitem[Silver et~al., 2017]{Silver2017}
Silver, D., Schrittwieser, J., Simonyan, K., Antonoglou, I., Huang, A., Guez,
  A., Hubert, T., Baker, L., Lai, M., Bolton, A., Chen, Y., Lillicrap, T., Hui,
  F., Sifre, L., van~den Driessche, G., Graepel, T., and Hassabis, D. (2017).
\newblock {Mastering the game of Go without human knowledge}.
\newblock {\em Nature}, 550(7676):354--359.

\bibitem[Spelke, 1990]{Spelke1990}
Spelke, E.~S. (1990).
\newblock {Principles of object perception}.
\newblock {\em Cognitive Science}, 14(1):29--56.

\bibitem[Spelke et~al., 1992]{Spelke1992}
Spelke, E.~S., Breinlinger, K., Macomber, J., and Jacobson, K. (1992).
\newblock {Origins of Knowledge}.
\newblock {\em Psychological Review}, 99(4):605--632.

\bibitem[Sutton and Barto, 2018]{SuttonBarto2018}
Sutton, R.~S. and Barto, A.~G. (2018).
\newblock {\em Reinforcement Learning: An Introduction}.
\newblock A Bradford Book, Cambridge, MA, USA.

\bibitem[Tsividis et~al., 2017]{Tsividis2017}
Tsividis, P.~A., Pouncy, T., Xu, J.~L., Tenenbaum, J.~B., and Gershman, S.~J.
  (2017).
\newblock {Human learning in Atari}.
\newblock In {\em 2017 AAAI Spring Symposium Series, Science of Intelligence:
  Computational Principles of Natural and Artificial Intelligence}.

\bibitem[Ulyanov, 2016]{Ulyanov2016}
Ulyanov, D. (2016).
\newblock Multicore-tsne.
\newblock \url{https://github.com/DmitryUlyanov/Multicore-TSNE}.

\bibitem[van~der Maaten and Hinton, 2008]{VanDerMaaten2008}
van~der Maaten, L. and Hinton, G. (2008).
\newblock Visualizing high-dimensional data using t-sne.

\bibitem[van Steenkiste et~al., 2018]{vanSteenkiste2018}
van Steenkiste, S., Chang, M., Greff, K., and Schmidhuber, J. (2018).
\newblock {Relational Neural Expectation Maximization: Unsupervised Discovery
  of Objects and their Interactions}.
\newblock In {\em International Conference on Learning Representations}.

\bibitem[Veerapaneni et~al., 2019]{Veerapaneni2019}
Veerapaneni, R., Co-Reyes, J.~D., Chang, M., Janner, M., Finn, C., Wu, J.,
  Tenenbaum, J.~B., and Levine, S. (2019).
\newblock {Entity Abstraction in Visual Model-Based Reinforcement Learning}.

\bibitem[Wang et~al., 2016]{Wang2016}
Wang, Z., Schaul, T., Hessel, M., {Van Hasselt}, H., Lanctot, M., and {De
  Frcitas}, N. (2016).
\newblock {Dueling Network Architectures for Deep Reinforcement Learning}.
\newblock In {\em 33rd International Conference on Machine Learning, ICML
  2016}, volume~4, pages 2939--2947. International Machine Learning Society
  (IMLS).

\end{thebibliography}

\FloatBarrier
\clearpage
\section{Appendix}

\begin{figure}[!htb]
\centering
\includegraphics[width=\linewidth]{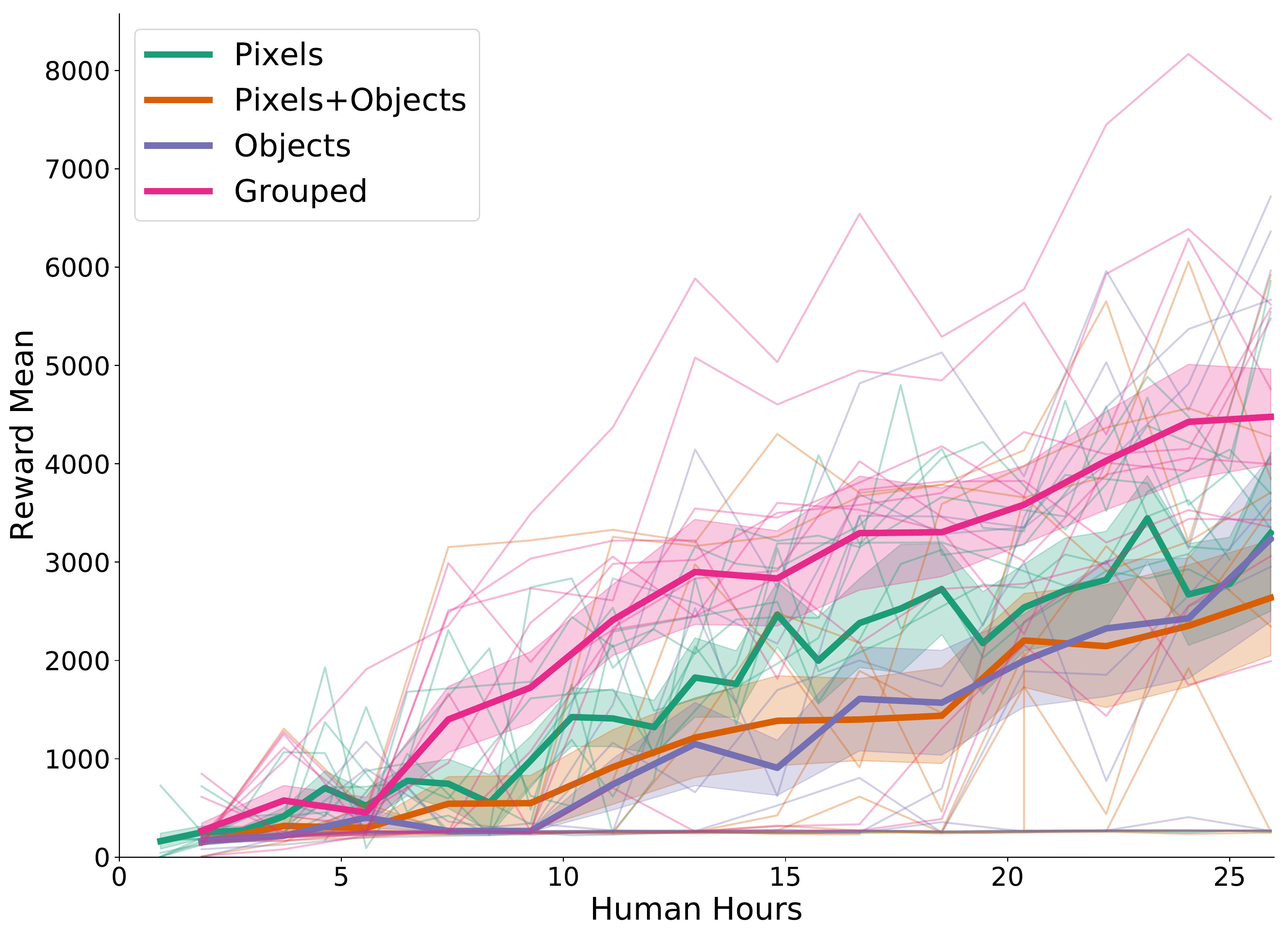}
\caption{ {\bf Mean evaluation reward in the first day of training.} 
Mean evaluation reward per condition, zoomed in on the first day or so (in human hours) of training. 
As in \autoref{fig:mean-results}, bold lines plot the mean per condition,  shaded areas reflect the standard error of the mean, and thin lines mark individual replications.}
\label{fig:mean-results-first-day}
\end{figure}

\end{document}